\pdfoutput=1

\documentclass[11pt]{article}

\usepackage[]{EMNLP2023}

\usepackage{times}
\usepackage{latexsym}

\usepackage[utf8]{inputenc}

\usepackage{microtype}

\usepackage{inconsolata}

\usepackage{booktabs}
\usepackage{multirow}
\usepackage{amsmath}
\usepackage{hyperref}
\usepackage{graphicx}

\usepackage{cleveref}

\usepackage{enumitem}
\setlist[description]{leftmargin=0cm, labelindent=0cm}

\usepackage[T4,T1]{fontenc}
\usepackage[verbose]{newunicodechar}
\newenvironment{tfour}{\fontencoding{T4}\selectfont}{}

\newcommand{\footurl}[1]{\footnote{\url{#1}}}

\title{Correcting FLORES Evaluation Dataset for Four African Languages}

\author{Idris Abdulmumin$^{1*+}$, Sthembiso Mkhwanazi$^2$, Mahlatse S. Mbooi$^2$, \\
    {\bf Shamsuddeen Hassan Muhammad$^{3*+}$, Ibrahim Said Ahmad$^{4*+}$, Neo Putini$^5$, }\\
    {\bf Miehleketo Mathebula$^1$, Matimba Shingange$^1$, Tajuddeen Gwadabe$^{*+}$, Vukosi Marivate$^{1,6}$}\\
    \footnotesize $^1$Data Science for Social Impact, University of Pretoria, $^2$Council for Scientific and Industrial Research, South Africa,\\
    \footnotesize $^3$Imperial College, London, $^4$Northeastern University, $^5$University of KwaZulu-Natal, $^6$Lelapa AI, $^{*}$HausaNLP, $^{+}$MasakhaneNLP \\
    \footnotesize correspondence: \texttt{idris.abdulmumin@up.ac.za}
}

\begin{document}
\maketitle
\begin{abstract}
This paper describes the corrections made to the FLORES evaluation (dev and devtest) dataset for four African languages, namely Hausa, Northern Sotho (Sepedi), Xitsonga, and isiZulu. The original dataset, though groundbreaking in its coverage of low-resource languages, exhibited various inconsistencies and inaccuracies in the reviewed languages that could potentially hinder the integrity of the evaluation of downstream tasks in natural language processing (NLP), especially machine translation. Through a meticulous review process by native speakers, several corrections were identified and implemented, improving the overall quality and reliability of the dataset. For each language, we provide a concise summary of the errors encountered and corrected and also present some statistical analysis that measures the difference between the existing and corrected datasets. We believe that our corrections improve the linguistic accuracy and reliability of the data and, thereby, contribute to a more effective evaluation of NLP tasks involving the four African languages. Finally, we recommend that future translation efforts, particularly in low-resource languages, prioritize the active involvement of native speakers at every stage of the process to ensure linguistic accuracy and cultural relevance.

\end{abstract}

\section{Introduction}

Low-resource languages, especially from Africa, are greatly under-represented in the Natural Language Processing (NLP) landscape, and this is primarily due to the absence of sufficient resources for both training and evaluation \cite{adelani-etal-2022-thousand,kreutzer-etal-2022-quality}. Various efforts have been made to create such resources and these include initiatives from organizations such as Lacuna\footurl{https://lacunafund.org/} that fund new and qualitative open datasets, and communities such as Masakhane, HausaNLP, the University of Pretoria's Data Science for Social Impact (DSFSI) Research Group, and other individual initiatives \cite{abdulmumin-etal-2022-hausa, parida-etal-2023-havqa}. For machine translation evaluation, the FLORES dataset \cite{flores101,nllb2022} is widely accepted as a benchmark for evaluation, especially because it was the first of its kind for many languages and enables many-to-many evaluation, making it easier to evaluate say a Hausa to Sepedi translation system without pivoting through a high resource language, e.g., English. Recently, the MAFAND dataset \cite{adelani-etal-2022-thousand} was created, but it only allows bilingual evaluation and is limited to the news domain.

While all these resources are being developed, it is imperative to review them for validation to ensure that they meet the expected standard of accuracy and representation.  
A revealing work by \citet{kreutzer-etal-2022-quality}, albeit on mostly web-crawled datasets, found that many of the datasets that are being relied upon for low-resource languages are littered with significant errors such as misalignments, incorrect translations, and other issues. The significance of evaluation datasets make them even more deserving of such reviews especially by literate native speakers that know how these languages are written and spoken. This paper, therefore, presents a comprehensive review and correction of the public FLORES evaluation datasets for four African languages: Hausa, Northern Sotho, Xitsonga and isiZulu. We also provide the corrected datasets for future evaluation tasks\footurl{https://github.com/dsfsi/flores-fix-4-africa}.

\section{The FLORES Evaluation Dataset}
The FLORES evaluation dataset consists of the first FLORES-101 \cite{flores101} and the subsequent more expanded FLORES-200 \cite{nllb2022} that included more languages.

\begin{description}
    \item[FLORES-101:] This was the original evaluation data and was created by translating the English dataset collected from Wikipedia, consisting of several topics and domains, into 101 mostly low-resource languages. The dataset was the first available evaluation benchmark for several low-resource languages and it enabled the evaluation of many-to-many translation systems without pivoting through another high-resource language such as English. Several quality control mechanisms were put in place to ensure that the final dataset was of acceptable quality. To determine if translations are good enough for inclusion in FLORES-101, a 20\% sample of the dataset were reviewed by language-specific evaluators who assess the quality using a Translation Quality Score (TQS) on a 0 to 100 scale, with a score of 90\% deemed acceptable. Errors such as grammar, punctuation, spelling, and mistranslation were examined, and each was assigned a severity level of minor, major, or critical. Three of the four languages in this paper were included in this dataset--Hausa (\texttt{hau}), Northern Sotho (\texttt{nso}) and Zulu (\texttt{zul}).
    \item[FLORES-200:] This dataset expanded FLORES-101 to over 200 languages, including our fourth target language--Xitsonga (\texttt{tso}). In this data, a more comprehensive process was developed to ensure the quality of the translations. Specifically, professional translators and reviewers aligned on language standards before the translators translated the sentences. Afterwards, automated checks were first conducted and then followed by manual checks by independent reviewers. Translations that were found lacking quality were sent back for post-editing. Similarly to FLORES-101, translations scoring above 90\% TQS were included in the FLORES-200.
\end{description}

\subsection{Problems Identified in FLORES}
Prior to this work, we have not found any published work that carefully reviews and attempt to correct mistakes in the FLORES evaluation dataset. However, some issues have been raised on the FLORES' public GitHub repositories.\footurl{https://github.com/openlanguagedata/flores} Some of these issues include near-identical translations in several dialects of Arabic: Mesopotamian (\texttt{acm}\textunderscore\texttt{arb}), Ta'izzi-Adeni (\texttt{acq}\textunderscore\texttt{arb}), Najdi (\texttt{ars}\textunderscore\texttt{arb}), and Moroccan (\texttt{ary}\textunderscore\texttt{arb}) Arabic dialects were found to be too similar to Standard Arabic (\texttt{arb}),\footurl{https://github.com/openlanguagedata/flores/issues/8}$^,$\footurl{https://github.com/facebookresearch/flores/issues/64} unspecifying the "orthography" and "variety" used in Lombard (\texttt{lmo}\textunderscore\texttt{latn}) and Sardinian (\texttt{srd}\textunderscore\texttt{latn}),\footurl{https://github.com/openlanguagedata/flores/issues/5}$^,$\footurl{https://github.com/openlanguagedata/flores/issues/6} unmatched quotation marks,\footurl{https://github.com/facebookresearch/flores/issues/36} and using Mandarin Chinese in Traditional Chinese Script (\texttt{zho}\textunderscore\texttt{Hant}) for Cantonese (\texttt{yue}\textunderscore\texttt{Hant}) translations.

\section{Focus Languages and Evaluation}

\subsection{Languages Covered}
In this work, the public\footurl{https://github.com/openlanguagedata/flores}  FLORES dev and devtest splits of Hausa, Northern Sotho (Sepedi), Xitsonga and isiZulu were reviewed and corrected by native speakers of the languages. A description of each language is presented in \Cref{sec:app_lang_desc}.

\subsection{Correction Guidelines}
For reviewing and subsequently correcting the identified errors in the datasets, the participants were given the following guidelines.

\begin{description}
    \item[Reviewing:] At this stage, the participants identified sentences in both data splits that require reviewing.
    \begin{itemize}
        \item \textbf{Read the original text:} carefully read the original English text to understand the intended meaning and context.
        \item \textbf{Compare with translated text:} compare each sentence or phrase in the original English text with its corresponding translation. Pay attention to both the overall meaning and the nuances of the language.
        \item \textbf{Check for accuracy:} look for errors, inaccuracies, or deviations from the original meaning in the translation. This includes mistranslations, omissions, additions, and grammatical mistakes.
        \item \textbf{Evaluate clarity and cohesion:} assess whether the translated text is clear and coherent in the target language. Ensure that it flows naturally and is easy for a target language-speaking audience to understand.
    \end{itemize}
    \item[Correcting the translations:] To correct the translations, we followed the guidelines provided in the shared task description.\footurl{https://oldi.org/guidelines\#translation-guidelines} The participants were trained on and encouraged to follow these guidelines when correcting the identified incorrect translations.
\end{description}

\begin{table*}[!ht]
    \centering
    \small
    \resizebox{\textwidth}{!}{
        \begin{tabular}{lrrrrrrrrrr}
        \toprule
           \multirow{2}{*}{lang} & \multicolumn{5}{c}{dev (997 sentences)} & \multicolumn{5}{c}{devtest (1,012 sentences)} \\\cmidrule(lr){2-6} \cmidrule(lr){7-11}
           & \# corr. (\%) & \# tokens$_o$ & \# tokens$_c$ & $\Delta$ tokens & \% div. & \# corr. (\%) & \# tokens$_o$ & \# tokens$_c$ & $\Delta$ tokens & \% div. \\
        \midrule
           \texttt{hau} & 632 (63.4) & 17,948 & 18,073 & 125 & 24.7 & 70 (6.9) & 2,006 & 1,978 & 28 & 49.2 \\
           \texttt{nso} & 67 (6.7) & 2,226 & 2,271 & 45 & 28.9 & 62 (6.1) & 2,082 & 2,105 & 23 & 28.0 \\
           \texttt{tso} & - & - & - & - & - & 83 (8.2) & 2,919 & 2,947 & 28 & 27.4 \\
           \texttt{zul} & 190 (19.1) & 3,605 & 3,588 & 17 & 23.7 & 226 (22.3) & 4,414 & 4,396 & 18 & 31.8 \\
        \bottomrule
        \end{tabular}
    }
    \caption{Data statistics; \# corr. (\%) $\rightarrow$ number of sentences requiring at least one correction (percentage of original data); \# tokens$_o\rightarrow$ original token count; \# tokens$_c\rightarrow$ corrected token count; $\Delta$ tokens $\rightarrow$ token count difference; \% div. $\rightarrow$ percentage of token divergence.}
    \label{tab:data_stats}
\end{table*}

\subsection{The Annotators}
The correction task was conducted by volunteer annotators that focused on their native languages. These annotators were a mix of university students and researchers holding first, second and third degrees in computing and linguistics.

\subsection{Evaluating the Corrections}
To determine the amounts of corrections and subsequent differences between the original and corrected data, we used the following metrics. The computations were conducted only on the instances that were corrected. We used the original dataset as the supposed predictions and for the reference translations, we used the corrected data. We used the Natural Language Toolkit (NLTK) \cite{bird-loper-2004-nltk} for all tokenization.

\begin{description}
    \item[Token Difference:] This is the difference between the number of all tokens in the original and corrected datasets.
    \item[Token Divergence:] This was used to measure the difference or dissimilarity between two sets of tokens. Given \(T_o\) and \(T_c\) as the set of tokens in the original and corrected datasets respectively, the following formula was used:
    \begin{equation}
        \text{divergence} = \frac{|T_o - T_c| + |T_c - To|}{|T_o \cup T_c|}
    \end{equation}
    The formula calculates the proportion of tokens that are different between the two sets relative to the total number of unique tokens across both texts. Higher divergence score indicates that the two texts are quite different, suggesting significant changes or corrections were made.
    
    \item[Translation Edit Rate:] \cite{snover-etal-2006-study} is a metric used in machine translation and other natural language processing tasks to measure the number of edits required to change a system-generated translation into a reference translation, and is computed using the following formula.
    \begin{equation}
        \text{TER} = \frac{\# \text{ of edits}}{\# \text{ of words in ref. translation}}
    \end{equation}
    The fewer the edits, the better the translation quality and a higher TER score indicates lower quality in the predicted translations.
    
    \item[BLEU:] \cite{papineni-etal-2002-bleu} is an n-gram based metric that indicates the quality of generated machine translations. The BLEU is computed as follows:
    \begin{equation}
        \text{BLEU} = BP \times \exp \left( \sum_{n=1}^{N} w_n \log p_n \right)
    \end{equation}
    where \(BP\) is the Brevity Penalty and is used to penalize instances where shorter translations are generated when the reference is comparably longer; \(p_n\) is the precision between the candidate translation and a set of ground truths; and \(w_n\) is the n-gram weights.
    
    \item[COMET:] \cite{rei-etal-2020-comet} is a metric that leverages pre-trained neural models and cross-lingual word embeddings to evaluate the quality of machine translation systems. We used the pre-trained models provided by \citet{wang-etal-2024-afrimte}.
\end{description}

\begin{table*}[!ht]
    \centering
    \small
    \begin{tabular}{lrrrrrrrr}
    \toprule
         \multirow{2}{*}{lang.} & \multicolumn{4}{c}{dev} & \multicolumn{4}{c}{devtest} \\
    \cmidrule(lr){2-5}\cmidrule(lr){6-9}
         & \multicolumn{2}{c}{TER} & \multirow{2}{*}{BLEU} & \multirow{2}{*}{COMET} & \multicolumn{2}{c}{TER} & \multirow{2}{*}{BLEU} & \multirow{2}{*}{COMET} \\
    \cmidrule(lr){2-3}\cmidrule(lr){6-7}
         & Score & \# Edits & & & Score & \# Edits & & \\
    \midrule
         \texttt{hau} & 19.2 & 3,107 & 72.0 & 54.1 & 40.4 & 711 & 56.6 & 42.1 \\
         \texttt{nso} & 22.4 & 472 & 68.5 & 55.2 & 21.2 & 409 & 71.8 & 55.9 \\
         \texttt{tso} & - & - & - & - & 20.9 & 547 & 73.9 & 58.4 \\
         \texttt{zul} & 17.2 & 524 & 76.3 & 53.0 & 23.6 & 879 & 70.6 & 53.0 \\
    \bottomrule
    \end{tabular}
    \caption{Similarities between the original and corrected FLORES evaluation data on the four African languages -- original as predictions; corrected as reference translations.}
    \label{tab:data_similarity}
\end{table*}

\section{Error Analysis}
\Cref{tab:data_stats,tab:data_similarity} present how similar, or different, the original sentences were to the corrections. Some of the errors found are analyzed below per language.

\paragraph{Hausa (\texttt{hau})}

A significant part of the translations were suspected to have been automatically generated, as many of them appeared incoherent or unclear. To investigate this, we conducted a comparison with translations from the Hausa FLORES dataset and new translations generated by Google Translate. The comparison revealed that, although there were limited exact matches\footnote{Google Translate may have evolved since the creation of the dataset.}, several incorrect lexical choices in the dataset's translations aligned with those produced by Google Translate, supporting the suspicion that the translations may have been automatically generated. It is important to note that other translation tools may exist for Hausa that we did not evaluate. Furthermore, several sentence-level translations from Google Translate were found to be more qualitative and coherent than those in the current dataset. For an illustration, we examine sentences from the dev and devtest sets, see \Cref{tab:hausa_examples}.

In several instance, named entities were translated instead of reusing them as they are due to the lack of their equivalents in Hausa. This is illustrated in the first example provided in \Cref{tab:hausa_examples}. Planned Parenthood appears as an organization that was not supposed to be translated (and may only be explained as \textit{hukuma mai kula da tsarin iyali}). The words in the organization name were translated as \textit{Iyayen Tsararru}, with their literal word translations (\textit{iyaye} $\rightarrow$ parents, \textit{tsararru} $\rightarrow$ planned) instead of the name of the organization as a named-entity. In the second example, the phrase "standard business attire" was translated as \textit{Kaya masu kala \begin{tfour}\m{d}\end{tfour}aya su ne cikakkun tufafin mu’amala} instead of \textit{kayan sawa na aiki da aka saba dasu}. The first translation is at best an incorrect explanation of the English phrase. And these are just two examples of the many we found in the dataset.

In addition to these severe mistakes, the dataset was littered with a lot of inconsistencies especially in the use of the standardized Hausa alphabets. Special characters are often ommitted and instead replaced with their normalized equivalents, e.g., \begin{tfour}\m{b}\end{tfour} $\rightarrow$ b, \begin{tfour}\m{d}\end{tfour} $\rightarrow$ d, etc. In some few places, the special \begin{tfour}\m{y}\end{tfour} is written as 'y which is acceptable.

\begin{table*}[]
    \centering
    \resizebox{\textwidth}{!}{
    \begin{tabular}{p{0.3cm}p{4.4cm}p{5.5cm}p{5.5cm}}
    \toprule
        \textbf{SN} & \textbf{English} & \textbf{Wrong Translation in FLORES} & \textbf{Corrected Translation} \\
        \midrule
        1. & Komen's policy disqualified Planned Parenthood due to a pending investigation on how Planned Parenthood spends and reports its money that is being conducted by Representative Cliff Stearns. & Manufar Komen ta hana \textcolor{red}{Iyayen Tsararru} sanadiyyar binciken kashe ku\textcolor{red}{d}i kan yadda \textcolor{red}{Tsararren Iyaye} yake ciyarwa kuma ta ba da rahoton \textcolor{red}{ku\begin{tfour}\m{d}\end{tfour}a\begin{tfour}\m{d}\end{tfour}\begin{tfour}\m{d}\end{tfour}inta} wanda Wakilin Cliff Stearns ke gudanarwa. & Manufar Komen ta dakatar da chanchantar \textcolor{blue}{Planned Parenthood} sanadiyyar binciken da akeyi akan yanda \textcolor{blue}{Planned Parenthood} take kashewa da kuma bayar da ba'asin ku\textcolor{blue}{\begin{tfour}\m{d}\end{tfour}}in ta wanda Wakili Cliff Stearns yake gudanarwa. \\ \midrule
        

        2. & Suits are standard business attire, and coworkers call each other by their family names or by job titles. & \textcolor{red}{Kaya masu kala \begin{tfour}\m{d}\end{tfour}aya su ne cikakkun tufafin mu’amala}, kuma abokan aiki kan kira junansu da sunan iyalinsu ko da mu\begin{tfour}\m{k}\end{tfour}am\textcolor{red}{a}n aiki. & Kwat sune \textcolor{blue}{kayan sawa na aiki da aka saba dasu} kuma abokan aiki suna kiran juna ne da sunan gidansu ko kuma matsayin da mutum yake kai.\\
        \bottomrule
    \end{tabular}
    }
    \caption{Some Hausa Examples of incorrect and inconsistent translations in FLORES dev and devtest.}
    \label{tab:hausa_examples}
\end{table*}

\paragraph{Northern Sotho (\texttt{nso})}

Several key challenges and areas for improvement were identified and corrected, focusing on vocabulary consistency, syntax, spelling, and the accurate conveyance of technical terms. Most of the text was accurately translated and, for the text with problems, only small changes were required to make it more accurate. Some of the words like “\textit{safatanaga} and \textit{disafatanaga}” have generally maintained lexical consistency although they were wrongly translated. These have been corrected to “\textit{sefatanaga} or \textit{difatanaga} (plural)”.

Although sometimes Sepedi uses borrowed words for many technical and scientific terms, things such as pavement do have a translation which could be “\textit{tsela ya maoto or tselanathoko}”. These could have been used instead of borrowing the pavement term to say \textit{pabamente}. The use of a borrowed term could have been from the available corpus or from learned behaviour for borrowing unknown English terms. Another example is the word college which was translated to \textit{colleje}, but Sepedi has a standard borrowed translation: “\textit{kholetšhe}”.

Addressing spelling errors and ensuring proper spacing between words are vital for readability and comprehension. For instance, the word "\textit{tswarelo}" was corrected to "\textit{tshwarelo}" to reflect the proper spelling. Similarly, "\textit{patlaladitše}" was adjusted to "\textit{phatlaladitše}", and "\textit{bontša}" to "\textit{bontšha}". Additionally, "\textit{mephutso}" should be spelt as "\textit{meputso}", and "\textit{delo}" should be corrected to "\textit{selo}". Spacing was required when using "\textit{begona}" so that it is "\textit{be gona}" and similar adjustments were made. These adjustments are crucial to maintain lexical consistency and to ensure that translations are accurate and easily understood.

Some terms were left out, like "scientific" as "\textit{tša bo ramahlale}" when scientific tools were talked about, and this greatly affected the meaning of the sentence. Additionally, in another instance, a sentence describing the use of Caesarean section to give birth to Nadia was misleading. Incorrectly, it implied that Nadia was both the baby being born and the individual undergoing the operation. This was corrected to have the intended meaning.

\paragraph{Xitsonga (\texttt{tso})}

Some of the problems identified in the Xitsonga translations included problems to do with vocabulary accuracy and the use of borrowed words. Among the errors that were identified is the translation of "Type 1 diabetes" to "\textit{vuvabyi bya chukela bya Type 1}". The correct phrase should therefore be "\textit{vuvabyi bya chukela bya muxaka wo sungula}", which captures the type of diabetes and avoid misunderstanding. Similar trends raise the importance of using proper terms that might fit local context as opposed to directly translating English words.

Another problem was that translations were mostly uniform, without contextual variations. Even here, the words "\textit{xiyenge xa \textcolor{red}{tlilinikhali na} sayense}" (clinical and scientific division) were used wrongly. The word actually is "\textit{xiyenge xa \textcolor{blue}{vutshila ni ntokoto bya} sayense}" (clinical and scientific division), but this clearly passes on the intended meaning. Moreover, the use of pluralization of terms was arbitrary. While the singular form of the term "worker" is "\textit{mutirhi}", the plural form should be "\textit{vatirhi}", and the singular form of "methods" is "\textit{maendlelo}", which should be in plural throughout instead of appearing in single forms.

Spelling problems and the usage of borrowed terms can have a substantial influence on the correctness of Xitsonga translations. One of the most illustrative examples of such incongruity of terms is that the English word "channel" has been translated as "\textit{chanele}". Instead, the work should have used the original term "\textit{nongonoko}" in order to ensure a perfect linguistic and connotative translation. To avoid generation of wrong impressions, the phrase borrowed from IsiZulu as used to mean "President" had to be replaced by the word "\textit{murhangeri wa tiko}" from Xitsonga. Deficient spelling, as in the case of writing "\textit{dokodela}" instead of "Dr", and examples of slang such as using "\textit{mwana wa}" instead of the formal "\textit{muongori}" indicate how borrowing and spelling mistakes reduced the quality of the translations. Fluency and correct spelling as well as using the native language correctly are a necessity to maintain the translated material’s effectiveness.

\paragraph{isiZulu (\texttt{zul})}

Similar to the errors identified in the other languages above, isiZulu translations displayed several common challenges. These included inconsistencies in vocabulary, syntax errors, and issues with the accurate expression of technical and scientific terms. The agglutinative nature of isiZulu and its conjunctive writing style further worsen these issues, leading to specific translation errors related to morphology and orthography.

A closer examination of these challenges reveals issues such as in the translation of "Around 11:29, the protest moved up Whitehall, ..." which was initially rendered as "\textit{Ngawo-11:29 ababhikishi baya Odongeni Olumhlophe, ...}". This translation contains two key issues. First, "\textit{Ngawo-11:29}" should have been "\textit{Ngabo-11:29}" to correctly match the grammatical structure for time expressions in isiZulu. Second, the literal transliteration of "Whitehall" as "\textit{Odongeni Olumhlophe}" failed to integrate properly into the sentence. The correct approach would involve incorporating the place name with the locative prefix "e-" to produce "e-Whitehall.". This prefix addition is required in conjunctive languages when using borrowed words or terms, but MT systems often fail to capture these variations. Additionally, another common issue was the unnecessary borrowing of words from English, despite the availability of standardized isiZulu terms. This was particularly evident with month names, scientific terms, and country names, where inconsistencies were frequent—one translation might use "January," another "\textit{uJanuwari}," and yet another "\textit{uMasingana}" Another example of this can be seen with the country name "Spain," which was inconsistently translated as both "Spain" and "\textit{Speyini}" in different sections. Similar inconsistencies occurred with attempts to translate organizational names or acronyms, leading to partial translations that disrupted the linguistic flow.

To address the inconsistencies, standardized isiZulu terms were consistently applied throughout the translations. For instance, month names such as "\textit{uMasingana}" replaced the inconsistent use of "January" and "\textit{uJanuwari}" In dealing with organizational names and acronyms and countries' names, the approach was to fully translate these entities or retain their original form consistently, avoiding partial translations that could disrupt the flow.

In addition to the inconsistencies with terminology, other errors were also identified and addressed. These included issues with verb conjugation, where incorrect tenses or forms were initially used, and the improper handling of borrowed words that did not align with isiZulu’s morphosyntactic rules. Minor spelling errors and incorrect use of prefixes or suffixes were also corrected to ensure that the translations were both grammatically accurate and easily understood.




\section{Conclusion}
In this work, we highlight the importance of qualitative evaluation datasets for low-resource languages and present our findings from a comprehensive review of the FLORES dataset for four African languages: Hausa, Northern Sotho, Xitsonga, and isiZulu. The original translations were marred by vocabulary inconsistencies, syntax errors, and inaccurate technical terms. After making necessary corrections, we measured the amount of edits and resulting difference between the improved datasets and the original using metrics like BLEU, TER, and COMET, which showed that significant improvements were made. The results presented highlight the need for ongoing refinement and human oversight in developing accurate translation datasets for underrepresented languages. For future work, we intend to expand the corrections to more African languages.

\bibliography{anthology,custom}
\bibliographystyle{acl_natbib}

\appendix

\section{Description of the Target Languages}
\label{sec:app_lang_desc}

\begin{description}
    \item[Hausa (\texttt{hau}):] Hausa is a widely spoken language across West Africa, particularly in Nigeria, Niger, Cameroon, and Ghana. It is spoken by approximately 77 million people worldwide, primarily in West Africa \cite{eberhard2022ethnologue}. Hausa ranks as the second most spoken language in Africa and 27th globally. The language belongs to the Chadic branch of the Afroasiatic language family, and it has a rich history of written communication. It was first written in Arabic script known as Ajami, reflecting the language's connection to Arabic, with many Hausa words borrowed from Arabic due to historical contact and influence. Today, the Boko script (also known as Roman script), which uses Latin characters, is the most common writing system for Hausa. This script excludes the letters p, q, v, and x, and includes additional consonants (\begin{tfour}\m{b}\end{tfour}, \begin{tfour}\m{d}\end{tfour}, \begin{tfour}\m{k}\end{tfour}, \begin{tfour}\m{y}\end{tfour}, kw, \begin{tfour}\m{k}\end{tfour}w, gw, ky, \begin{tfour}\m{k}\end{tfour}y, gy, sh, ts) and vowels (long a, i, o, u, e, and two diphthongs ai and au). Hausa follows a Subject-Verb-Object (SVO) sentence structure.

    \item[Northern Sotho (\texttt{nso}):]
    Northern Sotho, also known as Sepedi or Sesotho sa Leboa, is one of the official languages of South Africa and is spoken primarily by the Bapedi people in Limpopo Province. It is a Bantu language that belongs to the Sotho-Tswana group and shares linguistic similarities with Sesotho (Southern Sotho) and Setswana. Sepedi is known for its rich oral tradition that includes folklore, proverbs, and praise poetry that have played a significant role in the preservation of cultural heritage \cite{nurse2006bantu}.
    Sepedi is written using the Latin alphabet, with the standard 26 letters and a few additional characters such as the "š" which are adapted to its unique sounds. The language primarily follows a Subject-Verb-Object word order in sentence structure.
    
    \item[Xitsonga (\texttt{tso}):]
    Xitsonga, or Tsonga, is a Bantu language that is mainly spoken in South Africa and more especially in the Limpopo province and parts of Mpumalanga province. The language is estimated to be spoken by about 2.3 million people in South Africa.  Xitsonga belongs to the Niger-Congo language family, specifically the Tshwa-Ronga subgroup, and is characterized by the extensive use of prefixes and suffixes to convey meaning \cite{mabaso2018xitsonga}. This linguistic feature can impact the accuracy of translations, especially when dealing with technical and scientific concepts. It also feature a complex system of writing and syntax, which are prerequisites to clear and concise language usage. 
    Xitsonga is currently used in education and media section in South Africa, thus is regarded as relevant in cultural linguistic practices. That is why, the language being mentioned as a part of the country’s multiple languages system emphasizes its relevance and application in different phases of the people’s activity.

    \begin{table}[t]
        \centering
        \begin{tabular}{ll}
        \toprule
           Language & Sentence \\
        \midrule
           English & \textcolor{red}{I} \textcolor{blue}{know} \textcolor{purple}{them} \\
           Hausa & \textcolor{red}{Na} \textcolor{blue}{san} \textcolor{purple}{su} \\
           Northern Sotho & \textcolor{red}{Ndza} \textcolor{purple}{va} \textcolor{blue}{tiva} \\
           Xitsonga & \textcolor{red}{Ke} \textcolor{green}{a} \textcolor{purple}{ba} \textcolor{blue}{tseba} \\
           isiZulu & \textcolor{red}{Ngi}\textcolor{green}{ya}\textcolor{purple}{ba}\textcolor{blue}{zi}\\
        \bottomrule
        \end{tabular}
        \caption{The grammatical structure of different languages.}
        \label{tab:grammatical_structure}
    \end{table}
    
    \item[isiZulu \texttt{(zul}):] Zulu or isiZulu (in Zulu) is one of the 12 official languages in South Africa, and it is considered to be the most widely spoken indigenous language in the country. It constitutes approximately a quarter of the population, with around 15.1 million speakers out of the population of 62 million people \citep{statssa_statistics_2022}. IsiZulu is part of the Nguni language family, which is made up of a group of closely related Bantu languages belonging to a larger Niger-Congo language family, and they are widely spoken across Southern Africa \citep{mesham2021low}. These languages are particularly notable for their complex morphology, characterized by agglutinative morphology and conjunctive orthography. Agglutinative morphology means that words are typically formed by combining multiple small meaning-carrying units, known as morpheme. Conjunctive orthography  means that the morphemes are glued together to form a word, rather than writing them with spaces in between, as seen in disjunctive orthography, commonly associated with the Sotho group, as well as Tshivenda and Xitsonga in South Africa indigenous languages \citep{taljard2006comparison}. To illustrate this distinction, consider the example in Table \ref{tab:grammatical_structure} which examines the different grammatical structures of the phrase \textit{I know them}.

\end{description}



\Cref{tab:grammatical_structure} shows that while the phrase's meaning is consistent across languages, the writing systems vary: in disjunctive orthography, morphemes are separated by spaces, while in conjunctive orthography, as in isiZulu, they are joined into a single word. For example, in the phrase \textit{I know them}, each morpheme serves a specific grammatical function--`I' as the subject, `know' as the verb, and `them' as the object. In disjunctive orthography, these morphemes are written separately, making each unit distinct. In conjunctive orthography, they are combined into one continuous word, but the meaning remains intact. These orthographic variations pose challenges for machine translation systems, which must accurately process morphemes in different writing systems to produce accurate translations.

\end{document}